\documentclass[11pt]{article}

\usepackage[preprint]{acl}

\usepackage{times}
\usepackage{latexsym}
\usepackage{algorithm}
\usepackage{algpseudocode}
\usepackage{amsmath}
\usepackage{amsfonts}
\usepackage{pifont}
\usepackage{booktabs}
\usepackage{tabularx}
\usepackage{listings}
\usepackage{enumitem}
\usepackage{geometry}
\DeclareCaptionStyle{ruled}{labelfont=normalfont,labelsep=colon,strut=off} 
\usepackage[many]{tcolorbox}  
\newtcolorbox{boxL}{
    fontupper = \color{black},
    rounded corners,
    arc = 6pt,
    colframe = black!50, 
    boxrule = 0pt, 
    bottomrule = 4.5pt ,
    breakable,
}

\usepackage{listings}
\usepackage{xcolor}
\usepackage{placeins}
\lstdefinelanguage{json}{
    basicstyle=\ttfamily\small,
    numbers=left,
    numberstyle=\scriptsize,
    stepnumber=1,
    numbersep=6pt,
    showstringspaces=false,
    breaklines=true,
    rulecolor=\color{black!40},
    xleftmargin=1em,
    xrightmargin=1em,
    literate=
     *{:}{{:}}{1}
      {,}{{,}}{1}
      {"}{{"}}{1}
      {[}{{[}}{1}
      {]}{{]}}{1}
      {\{}{{\{}}{1}
      {\}}{{\}}}{1}
}

\usepackage{float}
\usepackage{subcaption}

\usepackage[T1]{fontenc}

\usepackage[utf8]{inputenc}

\usepackage{microtype}

\usepackage{inconsolata}

\usepackage{graphicx}

\title{Beyond Sentiment: A Multi-Agent Pipeline for Actionable Business Advice from Reviews}

\author{
Kartikey Singh Bhandari\textsuperscript{1} \quad Tanish Jain\textsuperscript{1} \quad Archit Agrawal\textsuperscript{1} \\
{\bf Dhruv Kumar\textsuperscript{1} \quad Praveen Kumar\textsuperscript{2} \quad Pratik Narang\textsuperscript{1}} \\
\textsuperscript{1}Birla Institute of Technology and Science, Pilani \\
\textsuperscript{2}Birdeye Inc. \\
\texttt{\{p20241006,f20230349,f20191048,dhruv.kumar,pratik.narang\}@pilani.bits-pilani.ac.in} \\
\texttt{praveen.kumar1@birdeye.com}
}

\begin{document}
\maketitle
\begin{abstract}
Customer reviews contain valuable signals about service quality, but converting large-scale review corpora into actionable business recommendations remains difficult. Standard sentiment/aspect analysis is largely descriptive, while direct prompting of large language models (LLMs) often yields generic and repetitive advice that is weakly grounded in user feedback. We propose a hierarchical decision-support pipeline that explicitly separates \textit{signal compression}, \textit{problem abstraction}, \textit{candidate generation}, \textit{objective-based evaluation}, and \textit{cost-aware routing} into different agents. This architectural decomposition produces auditable intermediate artifacts and enables controllable trade-offs between advice quality and token budget. Experiments on Yelp reviews from three service domains show consistent improvements over single-pass LLM baselines across multiple advice quality dimensions, including actionability, relevance, and non-redundancy. A human evaluation further indicates that users generally prefer our system’s recommendations. These results highlight the value of structured agentic decomposition for scalable, cost-aware business decision support.
\end{abstract}

\section{Introduction}

The vast operational potential of customer reviews remains largely untapped due to the limitations of traditional recommender systems in processing unstructured linguistic context \cite{Hasan2024ReviewbasedRS,wei2024llmrec,ren2024representation}. Although LLMs facilitate the extraction of sophisticated insights from such text \cite{zhang2024sentiment}, the prevailing focus in review mining has been on descriptive summaries rather than prescriptive outcomes. There is, therefore, a pressing need for frameworks that move beyond aspect extraction to provide actionable business recommendations.

Furthermore, reliance on a single LLM for the entire pipeline may constrain accuracy, factuality, and depth of reasoning \cite{huang2023towards,ji2023survey}. In response to this limitation, multi-agent LLM frameworks \cite{wang2024survey,10.24963/ijcai.2024/890,becker-etal-2025-mallm} have been proposed. For instance, \citet{wang2024macrec} introduced a collaborative framework in which specialized agents interact to generate customer-facing recommendations (e.g., what to buy or watch). However, such approaches \cite{hui2025matcha,zhang2024generative,zhang2024agentcf} are not designed to support managerial decision-making, where the goal is to determine what to fix.
\begin{figure*}[htbp]
\centering
\includegraphics[width=1.0\textwidth]{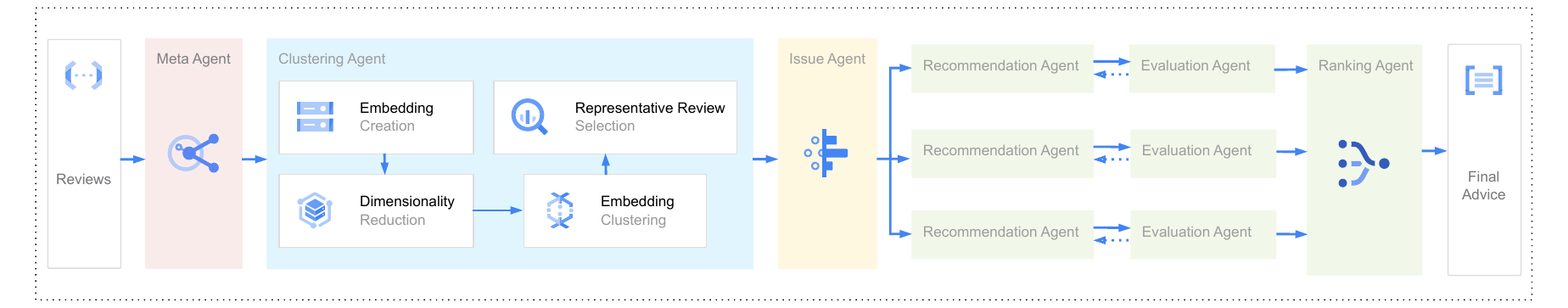}
\caption{\textit{\textbf{Overview of the proposed agentic decision-support pipeline.}} The Meta Agent first routes an incoming review corpus to an appropriate processing tier; in this figure it selects the full multi-agent configuration ($A_4$). The Clustering agent embeds and clusters reviews to select representative reviews, the Issue agent groups them into higher-level issues, Recommendation–Evaluation agents iteratively refine candidate interventions, and the Ranking agent prioritizes and outputs the final advice.}
\label{fig:Workflow}
\end{figure*}

A natural baseline is to prompt a single LLM to \textit{read the reviews and suggest improvements}. However, naive prompting is not a decision-support system: it tends to repeat redundant points across similar reviews, produce superficial, generic suggestions, overfit to the most salient or most recent examples rather than the full evidence distribution \cite{10.1162/tacl_a_00638}, lack an explicit mechanism to rank or select interventions under competing objectives, and ignore inference cost and thus cannot enforce a controllable quality-cost trade-off \citep{chen2024frugalgpt}.

In this paper, we propose a hierarchical decision-support pipeline for turning large-scale review corpora into actionable business guidance. Rather than treating advice generation as a single prompting step, our pipeline explicitly separates: 
\textit{signal compression} (clustering-based corpus distillation), 
\textit{problem abstraction} (issue modeling), 
\textit{candidate generation} (intervention proposals), 
\textit{objective-based evaluation} (iterative scoring and refinement), and 
\textit{cost-aware routing} (quality-cost control). 
This framing makes the contribution architectural where each module produces intermediate artifacts that are auditable and can be optimized independently. Section~\ref{sec:method} details each module and shows how the decomposition mitigates the failure modes of naive prompting. Our core contributions are as follows: 


\begin{enumerate}
\item We cast review mining as a prescriptive decision-support problem, selecting what to fix, rather than descriptive sentiment/aspect summarization.
\item We present a pipeline that distills reviews via clustering, abstracts them into issues, generates and selects interventions through objective-based iterative evaluation and explicit ranking under deployment oriented criteria.
 \item We develop a meta-agent that operationalizes user cost-quality preferences by routing execution to an appropriate configuration, enforcing the trade-off through system control flow.

\end{enumerate}
Together, these contributions provide a structured approach toward prescriptive, decision-relevant outputs that systematically address customer issues.


\section{Related Works}

\subsection{From Descriptive Analysis to Prescriptive Interventions}
Existing literature on review understanding has largely concentrated on descriptive tasks, most notably sentiment analysis, aspect extraction, and summarization \cite{zhang2024sentiment,tang2024aspect}. While recent LLM-based opinion summarization techniques have enhanced fluency and coverage, they remain primarily evaluative in nature. Parallel to this, review-aware recommender systems have integrated textual feedback into user-item modeling \cite{Hasan2024ReviewbasedRS,wei2024llmrec,ren2024representation}, with recent LLM-enhanced frameworks extending this by generating conversational justifications or natural language rationales \cite{wang2024macrec,hui2025matcha,zhang2024generative,zhang2024agentcf,gao2023chat,zhao2019personalized}. However, a critical limitation of these approaches is their optimization for consumer choice (determining \textit{what to select}) rather than provider improvement. \textit{In contrast to these lines of inquiry, our work addresses the underexplored prescriptive dimension of review analysis. We redirect language modeling capabilities toward managerial decision support, transforming descriptive feedback into prioritized, actionable interventions that address specific customer pain points.}
\subsection{Multi‑Agent and Ensemble LLM Frameworks}
Single LLM pipelines often lack factual depth and robustness \cite{huang2023towards,ji2023survey}, motivating multi‑agent architectures \cite{wang2024macrec,hui2025matcha,ye2025x,chen2023agentverse,yao2022react} that divide roles among generator, critic, and judge agents \cite{wang2024survey}. Such systems improve reliability through iterative critique, self-consistency, and ensemble selection. \textit{Our approach adapts this paradigm to review mining: clustering reviews to ensure coverage, iteratively refining advice for specificity, relevance, actionability, and concision and ranking outputs via a dedicated ranking agent. This aligns multi‑agent reasoning with business oriented prescriptive analytics rather than generic text generation.}

Complementary to multi-agent orchestration, recent work has studied cost-aware routing and cascades that select among LLMs to balance output quality against inference cost \cite{chen2024frugalgpt,ong2025routellm,dang2025multiagent,zhou-etal-2025-reso}. 
\textit{Motivated by this line of work, we introduce a meta-agent that routes each request to a pipeline configuration consistent with a user-specified cost-quality preference.}

\section{Proposed Method}
\label{sec:method}
Our method decomposes review-to-advice into five decision support primitives: 
We implement each primitive with a specialized agent.

\subsection{Clustering Agent}
\label{sec:signal_compression}

This agent performs \textit{signal compression}: it reduces a large, redundant review corpus into a compact evidence set that preserves coverage of major complaint modes while controlling downstream compute. 
Concretely, we cluster reviews in embedding space \cite{campello2013density,grootendorst2022bertopic}, rank clusters by support (cluster size), and select one representative review from each of the top-$m$ clusters. 
The output is a set of $m$ representative reviews that acts as a compressed proxy for the dominant issues in the full corpus.


To distill the corpus into a representative subset, we identify the review in each cluster that is most central to its semantic neighborhood. Let $\mathcal{D}=\{r_i\}_{i=1}^{N}$ denote the corpus of reviews, where each review $r_i$ is represented by a $d$-dimensional embedding $\mathbf{x}_i \in \mathbb{R}^d$ \cite{reimers2019sentence}. Given a cluster assignment $c(r_i) \in \{1, \dots, K\}$, we define $\mathcal{C}_k = \{r_i \in \mathcal{D} \mid c(r_i) = k\}$ as the set of reviews belonging to cluster $k$. The centroid $\boldsymbol{\mu}_k$ for each cluster is computed as the mean of its constituent embeddings:
\begin{equation}
\boldsymbol{\mu}_k = \frac{1}{|\mathcal{C}_k|} \sum_{r_j \in \mathcal{C}_k} \mathbf{x}_j.
\end{equation}

To extract a representative review $r_k^*$ for cluster $k$, we select the instance whose embedding maximizes the cosine similarity with the cluster centroid.

\begin{equation}
r_k^* = \arg\max_{r_i \in \mathcal{C}_k} \frac{\mathbf{x}_i^\top \boldsymbol{\mu}_k}{\|\mathbf{x}_i\|_2 \|\boldsymbol{\mu}_k\|_2}.
\end{equation}

\subsection{Issue Agent}
\label{sec:issue_modeling}
Although clustering yields representative reviews for dominant complaint modes, a single representative review may contain multiple issues, and the same minor issue can recur across representatives. 
To reduce redundant downstream computation and to improve coverage, we therefore extract all issue statements from the full set of representative reviews and consolidate them into a unified issue inventory. 
We then group and order these issues into higher-level themes, producing a structured, de-duplicated \textit{problem abstraction} that serves as the input to candidate generation. Appendix~\ref{sec:issue-output} provides examples of the issues-themes extracted.


\subsection{Recommendation Agent}
\label{sec:candidate_generation}

This agent performs \textit{candidate generation}: conditioned on the issue models, it proposes a diverse set of actionable interventions. 
The goal is not to produce a single \textit{best} answer in one pass, but to populate a candidate set that can later be evaluated and selected under explicit objectives.



\subsection{Evaluation Agent}
\label{sec:objective_eval}
Naive prompting lacks an explicit objective, so it cannot reliably discriminate between shallow and high-utility advice. 
We address this by introducing objective-based evaluation: each candidate intervention is scored on specificity (S), relevance (R), actionability (A), and concision (C), and the evaluator returns structured feedback used to iteratively refine the candidate \cite{shinn2023reflexion,madaan2023self}. 
Refinement stops when an explicit weighted objective exceeds a threshold or a certain number of iterations are reached.


\begin{equation}
w_{S} S + w_{R} R + w_{A} A + w_{C} C \ge 3.5
\end{equation}

where the weights $w_i$ are constrained such that:
\begin{equation}
\sum_{i \in \{S,R,A,C\}} w_i = 1, \quad w_i \in [0,1]
\end{equation}
Here, $w_{S}$, $w_{R}$, $w_{A}$, $w_{C}$ are set to default values of $0.25$ each. We set threshold as $3.5$ to require \textit{better than acceptable} quality, while limiting additional refinement iterations. We emphasize that this threshold is an operational design knob rather than a theoretically optimal bound, chosen to practically balance quality gains against compute cost. This converts advice generation into a constrained search process: generate candidates, score under an explicit utility function, and refine until the utility constraint is satisfied.

After this, we have a fully evaluated advice that is sent for further analysis. The following equations encompass the concepts used here.  
\begin{align}
r^{(t)} &= \mathcal{A}_{\text{Rec}}\!\left( f^{(t-1)} \right), \quad
f^{(t)} = \mathcal{A}_{\text{Eval}}\!\left( r^{(t)}, \;\boldsymbol{\theta} \right). \tag{3}
\end{align}

$\mathcal{A}_{\text{Rec}}$ denotes the recommendation agent, which generates a advice $r^{(t)}$ at iteration $t$, conditioned on feedback $f^{(t-1)}$ from the previous round.
$\mathcal{A}_{\text{Eval}}$ is the evaluation agent, which scores an advice against the evaluation parameters $\boldsymbol{\theta}$ and returns structured feedback $f^{(t)}$.
\begin{align}
r^{\ast} &= r^{(T)}, \quad 
T = \min \{\, t \mid f^{(t)} \models \boldsymbol{\theta} \;\lor\; t = T_{\max} \}. \tag{4}
\end{align}

The process continues iteratively until either the evaluation parameters pass the threshold or the maximum number of iterations $T_{\max}$ is reached. Here $T_{\max}$ act as a guard against infinite refinement \cite{bertsekas1999nonlinear}.

\subsection{Ranking Agent}
\label{sec:selection_ranking}
Decision support requires an explicit selection mechanism rather than returning the last generated output. 
We therefore generate multiple fully evaluated candidates (three in our implementation) and select among them, which improves robustness over single-pass generation \cite{wang2022self}. 
The ranking module uses an LLM judge to order candidates according to deployment-oriented criteria such as practicality, implementation cost, and expected efficacy \cite{zheng2023judging}. 
The top-ranked intervention is returned as the system output.

\subsection{Meta Agent}
\label{sec:meta_agent}
Our framework can be instantiated with multiple configurations that trade off
(i) advice quality and (ii) inference cost. As discussed in our cost-quality analysis
(\S\ref{sec:cost-quality}) and ablation results (Appendix \ref{sec:abl}), different agentic components and backbone choices
yield comparable mean quality in some settings while incurring materially different token budgets.
To provide a simple and transparent deployment control, we introduce a \emph{meta-agent} that
routes each user request to an appropriate configuration based on a user-specified preference
between accuracy and cost \cite{chen2024frugalgpt,ong2025routellm}. 
\paragraph{Preference score.}
Users provide nonnegative weights $(w_Q,w_C)$ for quality vs.\ cost. We convert these into a single
preference scalar \cite{miettinen1999nonlinear}
\begin{equation}
p \;=\; \frac{w_Q}{w_Q+w_C}\in[0,1],
\label{eq:pref_score}
\end{equation}
where larger $p$ indicates greater willingness to spend tokens for quality.

\paragraph{Routing over five pipelines.}
Let $\{A_0,\dots,A_4\}$ be five pipeline configurations ordered by increasing token cost
$C_0<\cdots<C_4$ (tokens-only). In our implementation: $A_0$ is a single-LLM baseline and
$A_1$--$A_4$ progressively add agentic components (see Appendix~\ref{app:meta_agent} for full
definitions and Table \ref{tab:example_costs} for costs). A standard utility-maximizing router can collapse intermediate tiers when
mean quality differences are small relative to token gaps. Since deployments often require
interpretable ``gears'' (stable, user-controllable tiers), we instead use an \emph{almost-uniform,
cost-aware discretization} of $p$. 

\begin{table}[ht!]
\centering
\small
\setlength{\tabcolsep}{6pt}
\begin{tabular}{c c c c c}
\toprule
$C_0$ & $C_1$ & $C_2$ & $C_3$ & $C_4$ \\
\midrule
4822 & 30045 & 30166 & 71737 & 114441 \\
\bottomrule
\end{tabular}
\caption{Token consumed for the five pipeline configurations for Automotive domain.}
\label{tab:example_costs}
\end{table}

The tokens consumed are similar in all three domains as the input is $m$ representative reviews for all domains and the pipelines are also same. Therfore, we calibrate routing thresholds on the Automotive split and apply them unchanged to Restaurant and Hospitality domains. 
\paragraph{Cost-aware, almost-uniform thresholds.}
We compute thresholds $b_i(\varepsilon)$ in a log-token geometry (Appendix~\ref{app:meta_agent}):
\begin{equation}
b_i(\varepsilon) \;=\; (1-\varepsilon)\frac{i}{5} \;+\; \varepsilon m_i,\quad i\in\{1,2,3,4\},
\label{eq:blended_thresholds_main}
\end{equation}
where $m_i$ are midpoints between adjacent normalized log-cost coordinates and
$\varepsilon\in[0,1]$ controls the trade-off between exact uniformity ($\varepsilon=0$) and
fully cost-aware binning ($\varepsilon=1$). We use a small $\varepsilon$ (e.g., $0.2$--$0.4$).

The meta-agent selects the tier index via
\begin{equation}
j \;=\; \sum_{i=1}^{4} \mathbf{1}[p \ge b_i(\varepsilon)],
\qquad A \leftarrow A_j,
\label{eq:routing_rule}
\end{equation}
which yields a monotone mapping from $p$ to exactly one of the five tiers.


\section{Experimentation}
The LLM used for issue generation is gpt-oss 20b \cite{openai2025gptoss} with a temperature of 0, while the LLM used for ranking is Gemini 2.5 Flash Lite \cite{google2025gemini25}.

We conducted a series of experiments to evaluate our framework across different model families and configurations. Throughout all experiments, we used \textit{all-MiniLM-L6-v2} to generate text embeddings. First, we applied the full framework to three large language models accessed through the Groq API (\textsc{Agentic-B}): \textit{Llama 3.3 70b Versatile} \cite{meta2025llama33}, \textit{GPT OSS 120b }\cite{openai2025gptoss}, and \textit{Deepseek R1 Distill Llama 70b} \cite{deepseek2025r1_70b}. We then evaluated the framework using three medium-sized models accessed via Ollama \footnote{\url{https://ollama.com/}} (\textsc{Agentic-M}):\textit{ Llama 3.1 8b} \cite{meta2024llama31}, \textit{Qwen 3 8b} \cite{qwen2025qwen3}, and \textit{Deepseek R1 8b} \cite{deepseek2025r1_8b}. We also tested the framework with three Gemini models, all based on\textit{ Gemini 2.5 Flash Lite} (\textsc{Agentic-G}). To establish baselines, we ran a simple one prompt, no framework pass using the same three large models separately:\textit{ Llama 3.3 70b Versatile} and \textit{Deepseek R1 Distill Llama 70b}; where the representative reviews are fed directly to these three large models to get the final advice.

We also conducted several ablation studies to assess the contribution of key components, including runs without the issue-generation agent, without the evaluation step, and without both (Appendix \ref{sec:abl}). The experimental setup information, details about metrics and quality analysis can be accessed from Appendix \ref{sec:exp}.

\begin{figure}[t]
\centering
\includegraphics[width=0.45\textwidth]{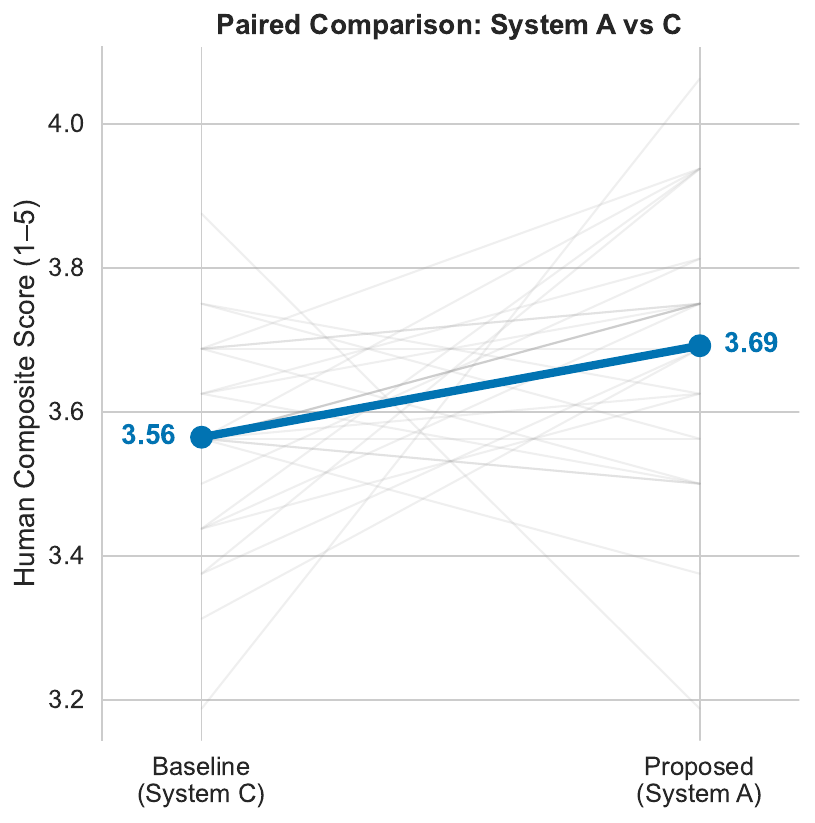}
\caption{Human evaluation (paired). Per-issue human composite scores (mean of 8 dimensions, averaged over two annotators) comparing the baseline (System C) and the proposed model (System A). Gray lines show paired scores for each issue; the bold line shows the mean.}
\label{fig:scatter}
\end{figure}

\subsection{Dataset}
The experimental corpus is sourced from the Yelp Open Dataset,\footnote{\url{https://www.yelp.com/dataset}} a large-scale repository of real-world service reviews. To ensure cross-domain robustness, we selected three distinct service industries representing diverse consumer interactions: Automotive (\textit{Payless Parking and Car Rental}), Hospitality (\textit{Circus Circus Reno}), and Restaurant (\textit{New Orleans Creole Cookery}). To facilitate a concentrated analysis of critical customer feedback and explicit service failures, we selectively sampled reviews with a 1-star rating. This filtering strategy prioritizes the extraction of high priority pain points, which are most prevalent in low rated feedback. The resulting corpus comprises $889$ negative instances; detailed distribution statistics relative to the total review volume for each business are summarized in Table \ref{tab:dataset-statistics}.

\section{Results}
\begin{table}[t]
\centering
\small
\setlength{\tabcolsep}{4pt}
\begin{tabular}{llrrr}
\toprule
SN. & Framework / Model      & Auto. & Res. & Hosp. \\
\midrule
1 & \textsc{Agentic-B}          & \textbf{95.2} & \textbf{94.7 }& 93.2 \\
2 & \textsc{Agentic-G}            & 94.7 & 94.7 & \textbf{94.4} \\
3 & \textsc{Agentic-M}       & 93.6 & 94.3 & 91.7 \\
4 & \texttt{deepseek-r1:70b}         & 94.1 & 92.2 & 91.6 \\
5 & \texttt{llama-3.3-70b}           & 93.1 & 89.8 & 86.5 \\
\bottomrule
\end{tabular}
\caption{Average composite advice quality scores (0–100) for multi-agent frameworks and single-model baselines across the automotive, restaurant, and hospitality domains.).}
\label{tab:framework-scores}
\end{table}
Across the three domains, the proposed multi-agent pipeline delivers consistently high composite recommendation quality across model families, with all three agentic configurations clustering tightly in the low-to-mid 90s and exhibiting only modest domain-dependent variation (Table \ref{tab:dataset-statistics}; Figure \ref{fig:experiments}). In particular, \textsc{Agentic-B} is strongest in automotive, while \textsc{Agentic-G} slightly leads in hospitality, suggesting that the scaffold is largely backbone-agnostic but can benefit from complementary model priors in more diverse service settings. By contrast, the single-pass baselines are noticeably more sensitive to backbone choice, with the weaker baseline degrading sharply in hospitality relative to the agentic variants. A consolidated per-dimension analysis relative to \textsc{Agentic-B} (Figure \ref{fig:heatmapexp}) further indicates that the observed gaps are driven primarily by losses in \emph{specificity} and \emph{non-redundancy} (and, secondarily, actionability/expected impact), whereas \emph{bias} and \emph{reading clarity} remain largely unchanged; interestingly, some non-agentic configurations appear more conservative on \emph{feasibility}, highlighting a trade-off where the multi-agent scaffold prioritizes evidence-grounded, structured interventions over overly cautious generic fixes.

\paragraph{Human evaluation.}
To sanity-check whether the LLM-as-judge conclusions are directionally consistent with expert preferences, we ran a blinded human study. Human ratings correlate positively with the judge at the composite level ($\rho=0.384$, $p=2\times10^{-4}$). Although inter-annotator agreement is modest (Appendix~\ref{app:human-eval}), this is expected because many rubric dimensions are highly skewed (most outputs receive 4-5), which makes $\kappa$ unstable despite high raw agreement. At the system level, humans still prefer the proposed method over the baseline under a paired Wilcoxon signed-rank test ($p=0.0097$). Finally, across issues treated as small ranking problems, the judge's Top-1 choice matches the human Top-1 choice on 20/30 issues (66.7\%).

\section{Discussion}


The most consistent qualitative effect of the scaffold is improved \emph{structure and traceability}. The clustering and issue identification stages provide a shared problem statement and a compact evidence representation, which helps downstream agents avoid repeatedly re-interpreting raw reviews \cite{li-etal-2025-generation} and converging on redundant boilerplate. This is reflected in the ablation behavior (Table \ref{tab:ablation}): removing the \textit{Issue Agent} causes the largest degradation, consistent with recommendations becoming more generic and less differentiated across issues, while removing the \textit{Evaluation Agent} mainly weakens dimensions that depend on iterative refinement (e.g., expected impact and novelty). The worst case occurs when both Issue and Evaluation are removed, indicating that these components contribute complementary value: the Issue Agent improves \emph{problem formulation and coverage}, whereas the Evaluation Agent improves \emph{solution quality control} through feedback-driven refinement \cite{shinn2023reflexion,madaan2023self}. Detailed per-domain scores and heatmaps are reported in the Appendix (see Appendix~\ref{sec:abl}).


\label{sec:cost-quality}
A central practical question for industry deployment is whether the quality gains from multi-agent orchestration justify the added inference cost \cite{chen2024frugalgpt,ong2025routellm}. Table~\ref{tab:example_costs} reports token consumption for five pipeline configurations in the automotive domain: the single-prompt baseline ($A_0$) costs $4,822$ tokens, while the full framework ($A_4$) increases cost to $114,441$ tokens, reflecting multiple agent calls and the refinement loop. The ablations clarify the cost structure. Configurations without the Evaluation Agent ($A_1$ and $A_2$) sit near $30$k tokens, whereas removing the Issue Agent ($A_3$) increases cost to $71,737$ tokens. This pattern reveals (i) iterative refinement contributes substantially to total cost when enabled ($A_4$ vs. $A_2$), and (ii) the Issue Agent acts as a compression step that reduces downstream processing ($A_2$ vs. $A_3$).

These patterns imply a clear deployment trade-off: when latency or budget is tight, practitioners can run a cheaper variant that skips iterative evaluation ($A_2$) while retaining most of the scaffold’s structural benefits; when quality is paramount, the full loop ($A_4$) provides stronger guardrails at higher cost.



\begin{table}[t]
\centering
\small
\setlength{\tabcolsep}{4pt}
\begin{tabular}{llrrr}
\toprule
SN. & Framework / Model                 & Auto. & Res. & Hosp. \\
\midrule
1 & Proposed Model                     & \textbf{95.2} & \textbf{94.7} & \textbf{93.2} \\
2 & w/o Issue Agent                    & 95.0 & 91.5 & 90.6 \\
3 & w/o Evaluation Agent               & 94.4 & 94.3 & 92.8 \\
4 & w/o Issue \& Evaluation Agent     & 95.2 & 90.3 & 91.0 \\
\bottomrule
\end{tabular}
\caption{Ablation study of the proposed multi-agent framework (\textsc{Agentic-B}), showing average composite advice quality scores (0–100) in the automotive, restaurant, and hospitality domains when removing the Issue Agent, the Evaluation Agent, or both.}
\label{tab:ablation}
\end{table}

\section{Conclusion}

We presented a multi-stage framework for generating actionable business advice from large-scale customer reviews. By decomposing the task into corpus distillation, issue abstraction, iterative refinement, and ranking, our approach reduces redundancy and improves grounding compared to single-pass prompting, while producing intermediate artifacts that support auditability. We further introduced a cost-aware routing that enables explicit trade-offs between recommendation quality and inference cost. Experiments across three Yelp domains show consistent improvements over prompt-based baselines, and human evaluation indicates that users generally prefer our recommendations. Overall, our results suggest that structured agentic decomposition can serve as a practical and scalable decision-support layer for review analytics, supporting real-world deployment under varying budget constraints with human oversight.

\section*{Ethical Considerations}
The proposed system functions strictly as a decision-support mechanism. A primary concern pertains to privacy; practitioners must redact personally identifiable information and minimize raw text exposure. Furthermore, it is important to acknowledge the inherent systematic biases in customer reviews. Although our clustering and traceability mechanisms mitigate over-amplification, outputs should not be interpreted as unbiased ground truth. Additionally, to preclude harmful recommendations stemming from model hallucinations, human review is strictly required prior to execution. Finally, recognizing the limitations of automated evaluation, we employ targeted human validation to substantiate our findings.



\bibliography{custom}
\clearpage
\newpage
\appendix

\section*{Appendix}
\label{sec:appendix}


\section{Limitations}
Although the framework demonstrates incremental gains in the caliber of advice provided, certain constraints provide avenues for further development. First, the depth of reasoning achieved by the full pipeline results in a higher computational footprint compared to single-pass baselines, suggesting a trade-off between absolute quality and inference speed \cite{chen2024frugalgpt,ong2025routellm}. Second, the experiments focused on a curated slice of highly negative feedback by sampling only 1-star reviews from a limited set of businesses and domains, which constrained conclusions about performance on mixed-sentiment inputs or broader customer populations. Finally, the inherent subjectivity in evaluating business actionability contributed to modest inter-rater agreement, underscoring the challenge of defining universal gold standards in prescriptive analytics \cite{li-etal-2025-generation}.
Future work will focus on narrowing the quality-cost gap. Because the full pipeline increases token usage significantly, we plan to explore caching and early-exit critique mechanisms to optimize inference efficiency without sacrificing output quality.

\section{Meta Agent Details}
\label{app:meta_agent}
Motivated by the cost-quality patterns in \S\ref{sec:cost-quality} and the differing contributions of
agentic components, we introduce a lightweight \emph{meta agent} that routes each
request to one of several pre-defined compute tiers based on a user-specified preference between
accuracy and cost.

\paragraph{User preference score.}
The user specifies nonnegative weights $(w_Q, w_C)$ indicating relative emphasis on quality
and cost. We convert these into a single scalar preference \cite{miettinen1999nonlinear}
\begin{equation}
p \;=\; \frac{w_Q}{w_Q+w_C} \in [0,1],
\label{eq:pref_score}
\end{equation}
where larger $p$ indicates greater willingness to spend tokens to improve output quality.

\paragraph{Candidate configurations and token costs.}
Let $\{A_0,\dots,A_4\}$ denote an ordered set of five pipeline configurations from cheapest to
most expensive, with measured token costs $C_0 < C_1 < C_2 < C_3 < C_4$ (tokens-only).
Concretely, we use:
\begin{itemize}
    \item $A_0$: Single-LLM inference (one-prompt, no framework),
    \item $A_1$: Framework w/o Issue Agent and w/o Evaluation Agent,
    \item $A_2$: Framework w/o Evaluation Agent,
    \item $A_3$: Framework w/o Issue Agent,
    \item $A_4$: Full multi-agent framework (Issue + Recommendation + Evaluation + Ranking).
\end{itemize}
This ordering matches the incremental inclusion of agentic components observed in the ablation
study (Appendix \ref{sec:abl}) and is aligned with the pipeline structure (Figure \ref{fig:Workflow}).  

\paragraph{Motivation: Why not use utility-optimal thresholds directly?}
A standard approach would select $A_j$ via an argmax of a scalarized utility
(e.g., $p\widetilde{Q}_j - (1-p)\widetilde{C}_j$). However, using mean quality scores alone can
cause intermediate configurations to become \emph{dominated} (never selected) when quality
differences are small relative to token costs. In practical deployments, teams often require
explicit, stable ``gears'' for budgeting, latency/cost envelopes, and operational policy
(including the ability to intentionally choose intermediate tiers). Therefore, we separate:
(i) preference elicitation (Eq.~\ref{eq:pref_score}) from (ii) a calibrated discretization that
ensures near-uniform coverage while still reflecting non-uniform token gaps.

\paragraph{Cost-aware uniformization via log-token geometry.}
Token costs are typically separated by multiplicative factors rather than additive increments.
We therefore define a normalized log-cost coordinate:
\begin{equation}
g_i \;=\; \frac{\log C_i - \log C_0}{\log C_4 - \log C_0},
\qquad i\in\{0,\dots,4\},
\label{eq:log_cost_coord}
\end{equation}
so that $g_0=0$ and $g_4=1$. Next, define cost-aware midpoints between adjacent configurations:
\begin{equation}
m_i \;=\; \frac{g_{i-1}+g_i}{2},
\qquad i\in\{1,\dots,4\}.
\label{eq:midpoints}
\end{equation}
If we used $\{m_i\}$ directly as thresholds, bins would be fully cost-aware but not uniform. For
a more interpretable control (approximately equal probability mass across the preference axis),
we blend these midpoints with uniform thresholds:
\begin{equation}
\begin{split}
u_i &= \frac{i}{5}, \quad i \in \{1,2,3,4\},\\
b_i(\varepsilon) &= (1-\varepsilon)\,u_i + \varepsilon\, m_i, \quad \varepsilon \in [0,1].
\end{split}
\label{eq:blended_thresholds}
\end{equation}

Here, $\varepsilon$ controls the trade-off between interpretability and cost-awareness:
$\varepsilon=0$ yields exactly uniform bins, while $\varepsilon=1$ yields purely cost-aware
(midpoint) bins. We use a small $\varepsilon$ (e.g., $0.2$--$0.4$) to obtain \emph{almost-uniform}
bins that still acknowledge large token jumps (notably from $A_0$ to $A_1$).

\paragraph{Routing rule.}
Let $b_0=0$ and $b_5=1$. The meta-agent selects the configuration index $j\in\{0,\dots,4\}$ by
\begin{equation}
j \;=\; \sum_{i=1}^{4} \mathbf{1}\!\left[p \ge b_i(\varepsilon)\right],
\qquad
A \leftarrow A_j,
\label{eq:routing_rule}
\end{equation}
which deterministically maps $p$ to exactly one of the five tiers.

\paragraph{Monotonicity and coverage.}
Assuming $C_0<\cdots<C_4$, the sequence $\{g_i\}$ is strictly increasing and so are the
midpoints $\{m_i\}$. Since $\{u_i\}$ is also strictly increasing, the blended thresholds
$\{b_i(\varepsilon)\}$ are strictly increasing for any $\varepsilon\in[0,1]$. Hence each interval
$[b_j, b_{j+1})$ has positive length, guaranteeing that every configuration $A_j$ is selected for
a non-empty range of $p$. This provides stable user control while remaining consistent with
token-cost geometry.

\paragraph{Example: token-only thresholds for the automotive setting.}
Using the measured token costs (tokens-only) in Table \ref{tab:example_costs}:

We compute $g_i$ from Eq.~\ref{eq:log_cost_coord}, then $m_i$ from Eq.~\ref{eq:midpoints}, and
finally the blended thresholds from Eq.~\ref{eq:blended_thresholds}. For $\varepsilon=0.3$,
this yields:
\begin{equation}
\left(b_1,b_2,b_3,b_4\right)
\approx
(0.2267,\;0.4535,\;0.6347,\;0.8379).
\label{eq:example_thresholds}
\end{equation}
Thus, the five routing bins are:
\[
\begin{aligned}
p &\in [0,0.2267) &\Rightarrow A_0, \\
p &\in [0.2267,0.4535) &\Rightarrow A_1, \\
p &\in [0.4535,0.6347) &\Rightarrow A_2, \\
p &\in [0.6347,0.8379) &\Rightarrow A_3, \\
p &\in [0.8379,1] &\Rightarrow A_4.
\end{aligned}
\]
These bins are close to uniform (nominally $0.2$ width) while reflecting the large multiplicative jump from $A_0$ to $A_1$ in token cost.
\paragraph{Backbone selection within tiers.}
Some tiers admit multiple backbone choices. We implement backbone selection as a second-stage choice that is
invoked only when needed:
\begin{equation}
M =
\begin{cases}
\{\textsc{Llama}, \textsc{DeepSeek}\}, & \text{if } A=A_0,\\
\textsc{Agentic-\{B, G, M}\}, & \text{if } A=A_4. 
\end{cases}
\label{eq:conditional_backbone}
\end{equation}
Our backbone choice is primarily driven by cost. Since the two single-LLM baselines consume a comparable number of tokens (and likewise the three multi-agent variants), we select backbones by comparing their OpenRouter per-token prices.
For the single-LLM baselines, \textit{Llama 3.3 70B Instruct} costs \$0.10/M input and \$0.32/M output tokens, whereas \textit{DeepSeek R1 Distill Llama 70B} costs \$0.03/M input and \$0.11/M output tokens.\footnote{\url{https://openrouter.ai/meta-llama/llama-3.3-70b-instruct}; \url{https://openrouter.ai/deepseek/deepseek-r1-distill-llama-70b}}
For the multi-agent variants, \textsc{AGENTIC-B} uses \{Llama 3.3 70B Instruct, gpt-oss-120b, DeepSeek R1 Distill Llama 70B\}, where gpt-oss-120b costs \$0.039/M input and \$0.19/M output tokens; thus AGENTIC-B averages \(\approx\$0.056/M\) input and \(\approx\$0.207/M\) output.\footnote{\url{https://openrouter.ai/openai/gpt-oss-120b/providers}}
\textsc{AGENTIC-G} uses three \textit{Gemini 2.5 Flash Lite} agents (so its per-agent and average price are the same): \$0.10/M input and \$0.40/M output.\footnote{\url{https://openrouter.ai/google/gemini-2.5-flash-lite}}
Finally, \textsc{AGENTIC-M} uses \{Llama 3.1 8B Instruct, Qwen3 8B, DeepSeek R1 0528 Qwen3 8B\}, priced at (\$0.02/\$0.05), (\$0.05/\$0.40), and (\$0.06/\$0.09) per million input/output tokens respectively, yielding an average of \(\approx\$0.043/M\) input and \(\approx\$0.180/M\) output.\footnote{\url{https://openrouter.ai/meta-llama/llama-3.1-8b-instruct}; \url{https://openrouter.ai/qwen/qwen3-8b}; \url{https://openrouter.ai/compare/deepseek/deepseek-r1-0528-qwen3-8b}}. The cost ranking is as follows:

\begin{itemize}
    \item \textbf{Single-LLM baselines:} 
    $\textit{Llama 3.3 70B} > \textit{DeepSeek R1 Distill 70B}$
    \item \textbf{Multi-agent variants:} 
    $\textsc{AGENTIC-G} > \textsc{AGENTIC-B} > \textsc{AGENTIC-M}$
\end{itemize}

\paragraph{Discussion.}
The proposed meta-agent provides a simple interface for practitioners to trade off inference cost
and output quality while maintaining interpretable, stable compute tiers. By blending uniform and
cost-aware thresholds, the routing policy yields near-uniform preference bins yet respects the
fact that token budgets are not evenly spaced across pipeline variants. This design is particularly
useful in settings where mean quality differences between variants are modest (as observed in our
cost--quality results) but operational policies require explicit intermediate budget options.

\section{Experimentation}\label{sec:exp}
We select representative reviews from the top $m$ clusters (by cluster size) to balance coverage with downstream inference cost; unless otherwise stated, we set $m=5$ in all experiments as a practical default that yields sufficient thematic diversity while keeping the multi-agent pipeline tractable under serving constraints.

\begin{table}[t]
\centering
\small
\setlength{\tabcolsep}{6pt}
\begin{tabular}{llrrr}
\toprule
SN. & Domain & Total Count & 1-Star & \% of Total \\
\midrule
1 & Automotive & 403 & 351 & 87.1 \\
2 & Hospitality & 1,292 & 365 & 28.3 \\
3 & Restaurant & 1,252 & 173 & 13.8 \\
\midrule
& \textbf{Total} & \textbf{2,947} & \textbf{889} & \textbf{30.2} \\
\bottomrule
\end{tabular}
\caption{Dataset: Distribution of 1-star reviews across the three selected business domains from the Yelp Open Dataset.}
\label{tab:dataset-statistics}
\end{table}
\subsection{Setup} 
All experiments were conducted on a workstation equipped with an Intel Xeon Silver 4310 CPU @ 2.10GHz, 128 GB RAM, and 3× NVIDIA A6000 GPUs with 48 GB memory each. The software environment consisted of Python 3.10, PyTorch 2.1, and CUDA 12.0.
\subsection{Quality Analysis}
To evaluate the generated business advice, we employ an LLM-as-a-judge framework \cite{zheng2023judging}. This approach facilitates consistent scoring across various models and recommendation formats. We use \textit{Llama-3.1-8B} \footnote{\url{https://huggingface.co/meta-llama/Llama-3.1-8B}} as the evaluator, configured with a temperature of 0.1 to ensure reproducibility and minimize stochasticity. The judge model assesses each recommendation against the eight dimensions defined below, with all final scores derived from the model's structured ratings.
\begin{tcolorbox}[title={Prompt Structure},colframe=black!30!gray]
You are an expert evaluator assessing business recommendations.
Rate the following recommendation on all 8 quality dimensions.
\begin{lstlisting}[language=json, numbers=none, xleftmargin=0pt]
    
CONTEXT:
Business Type: {business_context}
Original Reviews/Issues: {original_context}

RECOMMENDATION TO EVALUATE:
{recommendation}

DIMENSIONS TO RATE (1-5 scale for each):
1. Actionability: [definition and scale]
2. Specificity: [definition and scale]
...

RESPONSE FORMAT (JSON only):
{
    "actionability": <integer 1-5>,
    "specificity": <integer 1-5>,
    ...
}
\end{lstlisting}    

\end{tcolorbox}
\subsubsection{Metrics}
\label{sec:metrics}

Each advice is evaluated based on the following criteria:

\begin{enumerate} \item \textbf{Actionability:} Quantifies the degree to which a recommendation prescribes concrete, operational steps. High scores are assigned to suggestions with clear ownership and time-bound procedures, whereas lower scores are indicative of vague or purely aspirational statements.

\item \textbf{Specificity:} Assesses the level of granular detail (e.g., who, where, when) provided within the recommendation. This metric evaluates the extent to which advice is empirically grounded in issues rather than relying on generic cliches.

\item \textbf{Feasibility:} Evaluates the practical viability of the recommendation within the typical resource constraints of small-to-medium businesses (SMBs), including financial costs, personnel expertise, and implementation effort.

\item \textbf{Expected Impact:} Measures the projected efficacy of the recommendation in influencing key performance indicators (KPIs), such as customer retention or operational efficiency. Scores reflect the degree to which the suggested intervention is material to the identified issues.

\item \textbf{Novelty:} Examines the extent to which the model generates non-trivial insights that move beyond obvious "hygiene factors." This dimension prioritizes original, data-driven opportunities over commonplace or generic advice.

\item \textbf{Non-redundancy:} Evaluates the synthetic quality of the output, rewarding the integration of multiple signals into a concise format. This prevents the repetitive paraphrasing of source content in favor of prioritized information density.

\item \textbf{Bias:} Assesses the neutrality and objectivity of the text. High-scoring recommendations must be free from unfounded assumptions, stereotypes, or problematic biases, remaining strictly evidence-based.

\item \textbf{Reading Clarity:} Captures the linguistic accessibility, coherence, and professional tone of the recommendation, ensuring the output is grammatically sound and easy to interpret.
\end{enumerate}
\subsubsection{Scoring and Normalization}\label{sec:scoring}Each dimension is initially rated by evaluators on a 5-point Likert scale. To facilitate a more granular comparison, these raw ratings are linearly transformed into a normalized range of $[0, 100]$. Formally, for a raw score $x \in \{1, \dots, 5\}$, the scaled score $s_{\text{scaled}}$ is calculated as follows:\begin{equation}s_{\text{scaled}} = 100 \times \frac{x - 1}{4}\end{equation}

The mapping is as follows: $1$: Poor; $2$: Below Average; $3$: Average; $4$: Good; $5$: Excellent. Finally, a Composite Quality Score is derived by calculating the arithmetic mean of all eight dimension scores, providing a singular holistic metric for model performance.

\begin{figure*}[h]
\centering
\includegraphics[width=1\textwidth]{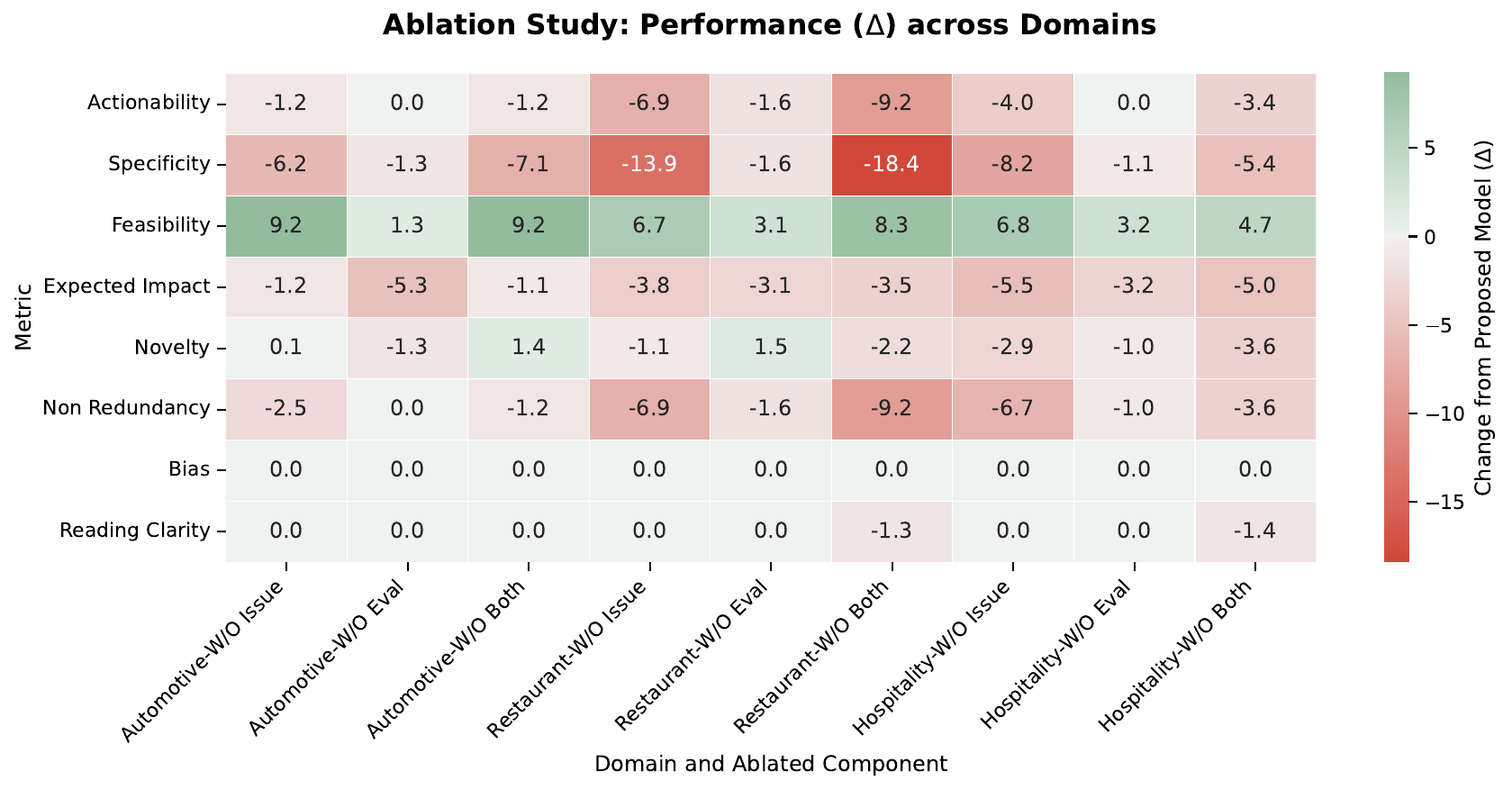}
\caption{Ablation impact by domain and metric ($\Delta$ from proposed model).Heatmap of the change in automatic evaluation score (percentage points) when removing components of the multi-agent pipeline: Issue Agent, Evaluation Agent, or both, across three domains (automotive, restaurant, hospitality). Each cell reports the score difference relative to the full proposed system (green = higher, red = lower).
}
\label{fig:Ablation}
\end{figure*}

\begin{figure*}[h]
\centering
\includegraphics[width=1\textwidth]{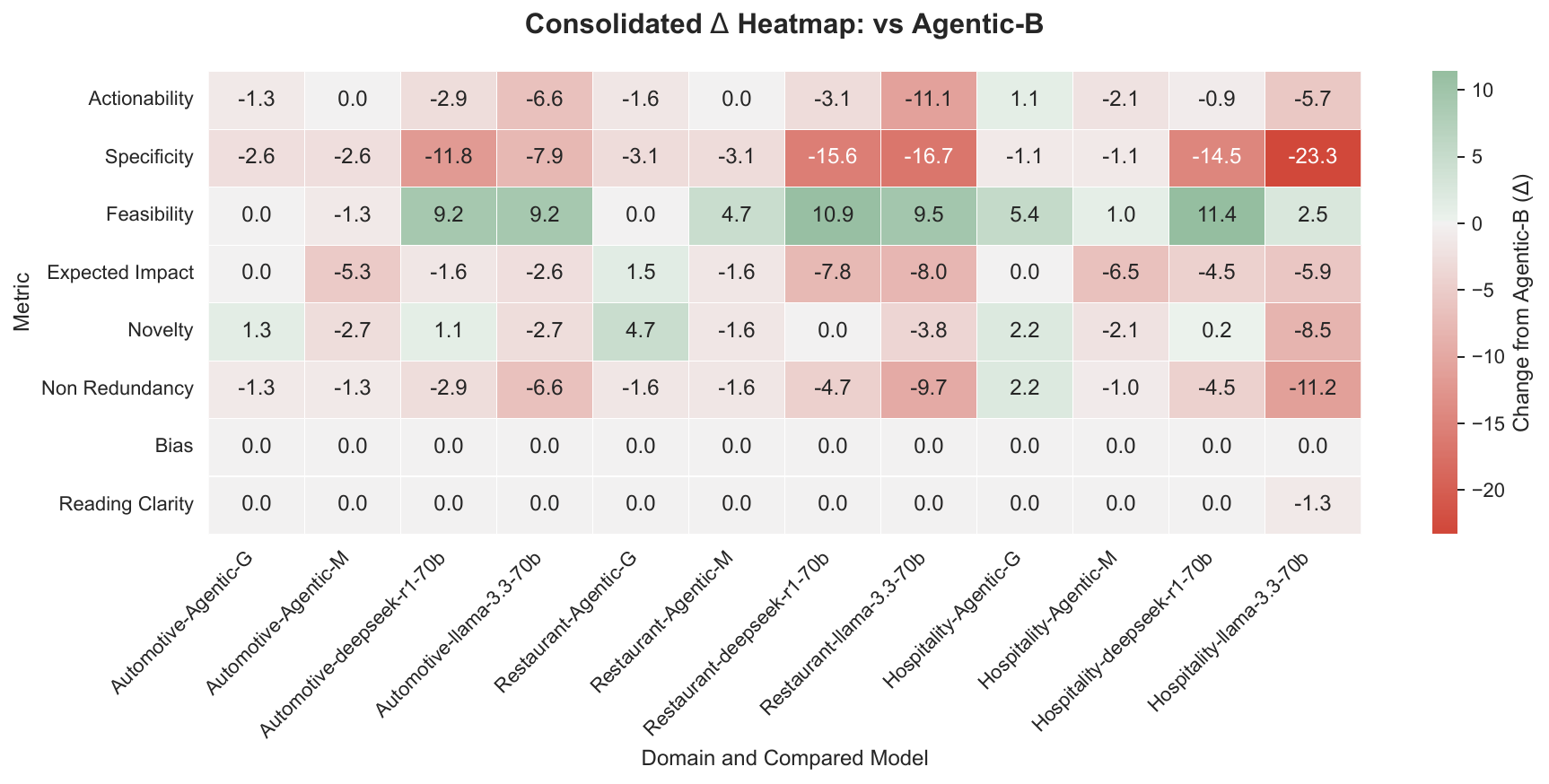}
\caption{Consolidated per-metric comparison against Agentic-B ($\Delta$). Heatmap of score differences between each compared system and Agentic-B across three domains (automotive, restaurant, hospitality) and eight metrics. Each cell reports the change relative to Agentic-B ($\Delta$ = compared model - Agentic-B; green = higher, red = lower).}
\label{fig:heatmapexp}
\end{figure*}
\section{Ablation Study}
\label{sec:abl}
\subsection{Results}
Across all three domains, the full multi-agent configuration achieves the highest or tied-highest composite quality scores: 95.2 for automotive, 94.7 for restaurant, and 93.2 for hospitality (0–100 scale).Ablating the \emph{Issue Agent} consistently degrades performance, with the largest drops in Restaurant (94.7$\rightarrow$91.5, $-3.2$) and Hospitality (93.2$\rightarrow$90.6, $-2.6$), while Automotive is only marginally affected (95.2$\rightarrow$95.0, $-0.2$). Removing the \emph{Evaluation Agent} yields smaller, more uniform reductions (Auto. $-0.8$; Res. $-0.4$; Hosp. $-0.4$), suggesting that single-pass generation remains competitive on average quality but benefits from iterative critique. The strongest degradation occurs when \emph{both} Issue and Evaluation Agents are removed, especially in Restaurant (94.7$\rightarrow$90.3, $-4.4$) and Hospitality (93.2$\rightarrow$91.0, $-2.2$). Notably, Automotive remains tied under joint removal (95.2), indicating domain-dependent sensitivity.

\subsection{Discussion}
Overall, the ablations support complementary roles for the agents. The \emph{Issue Agent} appears most critical in domains where multi-factor complaints require consolidation and de-duplication, producing more targeted and less repetitive recommendations. The \emph{Evaluation Agent} provides refinement pressure that improves quality more modestly but consistently across domains, plausibly by filtering weaker suggestions and tightening the rationale for high-impact actions. The larger sensitivity in Restaurant/Hospitality suggests these settings benefit most from explicit issue structuring and iterative critique, whereas Automotive advice may be more robust to component removal at the composite level.

\section{Human Evaluation Details}
\label{app:human-eval}

Our main experiments rely on LLM-as-judge scoring for scalability. To assess whether these automatic scores are directionally consistent with human judgments, we conducted a blinded human evaluation on a stratified subset of outputs and quantify (i) inter-annotator agreement, (ii) human--judge correlation, (iii) rank agreement, and (iv) statistical significance of system-level differences.

\paragraph{Protocol and data.}
We sampled 30 issues in total (10 per domain) and evaluated three systems, proposed model (\textsc{System A}) , proposed model w/o evaluation agent (\textsc{System B}) and baseline (Llama 3.3 70B) (\textsc{System C}) per issue (90 outputs). Each output was rated independently by two human experts on our eight quality dimensions using a 1--5 Likert scale (Actionability, Specificity, Feasibility, Expected Impact, Novelty, Non-redundancy, Bias, Reading Clarity). Following the main paper, we compute a \emph{human composite} as the arithmetic mean across the eight dimensions (range 1--5). For comparability with the LLM judge, we also report judge composites computed as the arithmetic mean of per-dimension scores after rescaling 1--5 ratings to a 0--100 range (Appendix \ref{sec:scoring}). 

\paragraph{Inter-annotator agreement (IAA).}
Table~\ref{tab:human-iaa} reports quadratic-weighted Cohen’s $\kappa$ (ordinal) per dimension and the average across dimensions. Agreement is modest overall ($\overline{\kappa}=0.197$), with higher agreement for Feasibility ($0.533$) and Expected Impact ($0.358$), and near-zero agreement for some dimensions (e.g., Non-redundancy, Bias, Reading Clarity) due to limited variance in expert ratings. Notably, $\kappa$ is lower for skewed dimensions.

\begin{table}[t]
\centering
\small
\begin{tabular}{l c}
\toprule
\textbf{Metric} & \textbf{Quadratic $\kappa$} \\
\midrule
Actionability & 0.104 \\
Specificity & 0.376 \\
Feasibility & 0.533 \\
Expected Impact & 0.358 \\
Novelty & 0.250 \\
Non-redundancy & -0.004 \\
Bias & 0.000 \\
Reading Clarity & -0.044 \\
\midrule
\textbf{Average} & \textbf{0.197} \\
\bottomrule
\end{tabular}
\caption{Human--human agreement on the 30-issue subset (quadratic-weighted Cohen’s $\kappa$; ordinal 1--5 scale).}
\label{tab:human-iaa}
\end{table}

\begin{figure}[h]
\centering
\includegraphics[width=0.92\linewidth]{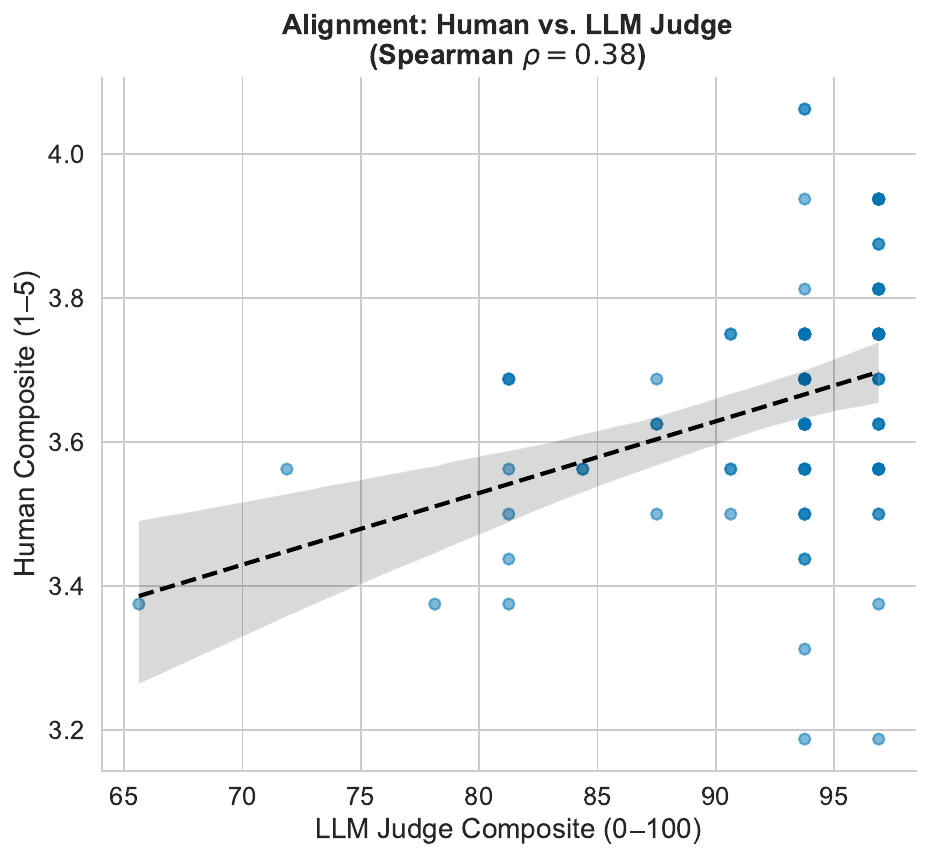}
\caption{Human vs. LLM-judge composite alignment
on the 30-issue subset (each point is one system output).}
\label{fig:human-judge-scatter}
\end{figure}

\paragraph{Human--judge alignment.}
Figure~\ref{fig:human-judge-scatter} plots human composite scores against LLM-judge composite scores. We observe a positive and statistically significant association (Spearman $\rho=0.384$, $p=0.0002$), suggesting that the automatic judge is broadly directionally aligned with human preferences at the composite level. Table~\ref{tab:human-judge-corr} provides per-metric Spearman correlations; results for Bias and Reading Clarity are marked as N/A because the automated judge produced constant scores (100/100) for all outputs in this subset, resulting in zero variance.


\begin{table}[h]
\centering
\small
\begin{tabular}{l c}
\toprule
\textbf{Metric} & \textbf{Spearman $\rho$ (Human vs Judge)} \\
\midrule
Actionability & 0.330 \\
Specificity & -0.005 \\
Feasibility & 0.360 \\
Expected Impact & 0.399 \\
Novelty & 0.107 \\
Non-redundancy & 0.184 \\
Bias & N/A (constant judge scores) \\
Reading Clarity & N/A (constant judge scores) \\
\midrule
\textbf{Composite} & \textbf{0.384} \\
\bottomrule
\end{tabular}
\caption{Human--judge correlation by metric on the 30-issue subset.}
\label{tab:human-judge-corr}
\end{table}

\paragraph{Rank agreement (Top-1).}
For each issue, we can treat the three system outputs as a small ranking problem. The LLM judge’s top-1 system matches the human top-1 system for \textbf{20/30} issues (\textbf{66.7\%}). This rank-level metric is often more stable than pointwise correlations for subjective rubrics.

\paragraph{System-level comparison on human scores.}
Table~\ref{tab:human-sig} reports mean human composite scores across the 30 issues and paired Wilcoxon signed-rank tests (per-issue pairing). Both the proposed system (System A) and the alternative multi-agent variant (System B) significantly outperform the vanilla baseline (System C), while the two multi-agent variants are not statistically distinguishable on this subset.

\begin{table}[h]
\centering
\small
\begin{tabular}{l c}
\toprule
\textbf{System} & \textbf{Mean human composite (1--5)} \\
\midrule
System A (Proposed) & 3.692 \\
System B (Ablation) & 3.708 \\
System C (Baseline) & 3.565 \\
\bottomrule
\end{tabular}
\caption{Mean human composite scores on the 30-issue subset.}
\label{tab:human-means}
\end{table}

\begin{table}[h]
\centering
\small
\begin{tabular}{l c c}
\toprule
\textbf{Comparison} & \textbf{$p$-value} & \textbf{Test} \\
\midrule
A vs C & 0.0097 & Wilcoxon signed-rank \\
A vs B & 0.5879 & Wilcoxon signed-rank \\
B vs C & 0.0011 & Wilcoxon signed-rank \\
\bottomrule
\end{tabular}
\caption{Paired significance tests on per-issue human composite scores ($N=30$).}
\label{tab:human-sig}
\end{table}

\begin{figure}[h]
\centering
\includegraphics[width=0.90\linewidth]{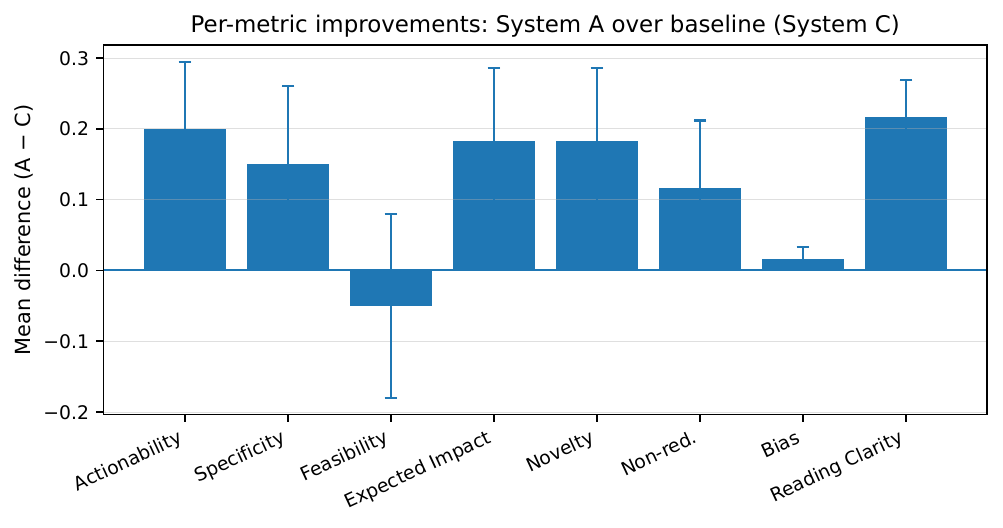}
\caption{Per-metric human-rated improvement of System A over System C (mean difference with standard error).}
\label{fig:human-metric-delta}
\end{figure}

\paragraph{Qualitative visualization.}
Figure~\ref{fig:scatter} visualizes per-issue paired differences between System A and System C; most issues show an improvement for System A. Figure~\ref{fig:human-metric-delta} shows the average per-metric improvement (A$-$C) with standard error bars, highlighting which rubric dimensions contribute most to the human-rated gains. 


\begin{figure}[h]
\centering
\includegraphics[width=0.5\textwidth]{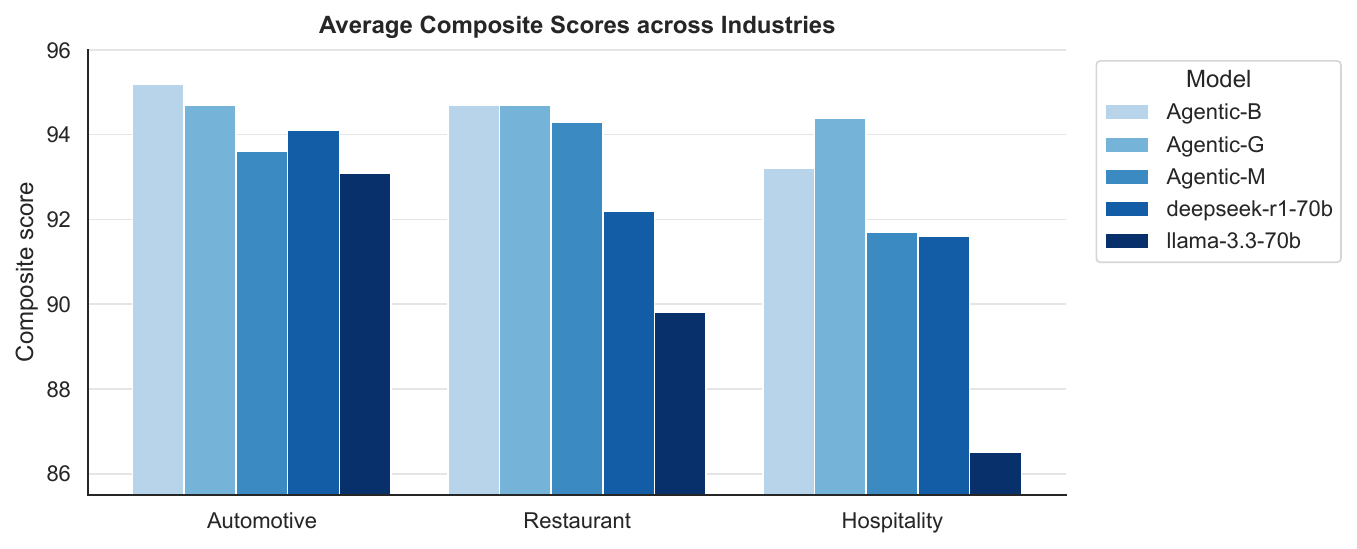}
\caption{Average composite quality by domain. Mean composite evaluation scores (averaged over the eight rubric dimensions) for each compared model across the three domains (automotive, restaurant, hospitality).}
\label{fig:experiments}
\end{figure}

\begin{figure}[h]
\centering
\includegraphics[width=0.5\textwidth]{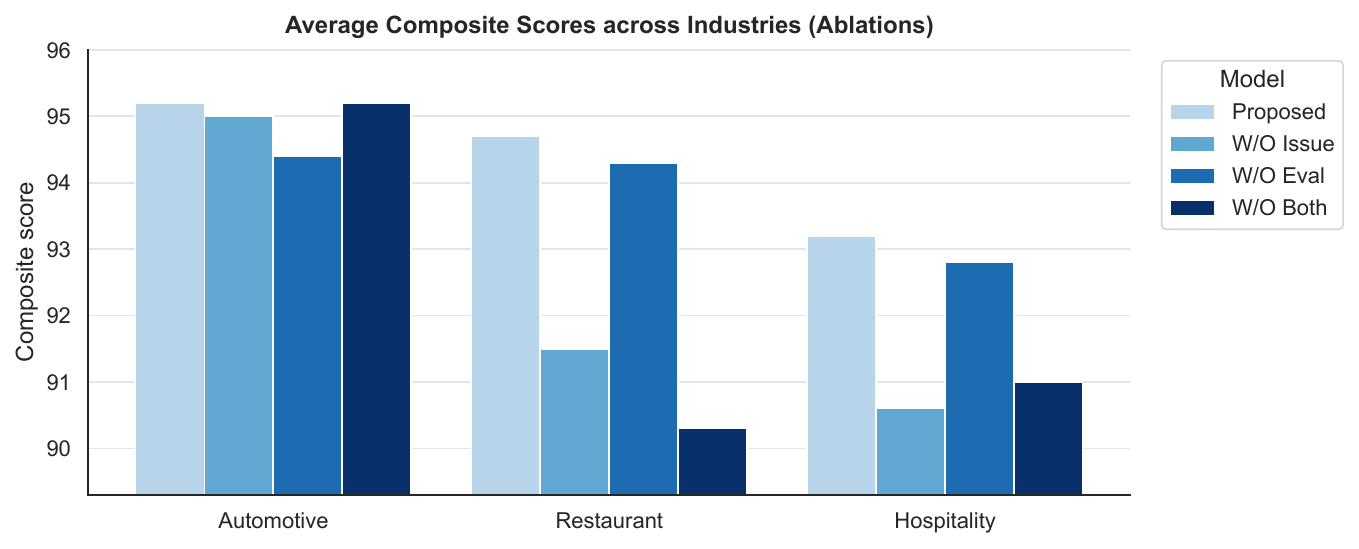}
\caption{Ablation effects on composite quality across domains. Mean composite scores (averaged over eight dimensions) for the full proposed pipeline and ablated variants removing the Issue agent (w/o Issue), the Evaluation agent (w/o Eval), or both (w/o Both), reported separately for automotive, restaurant, and hospitality.}
\label{fig:ablresults}
\end{figure}



\clearpage
\onecolumn
\section{Prompts}
\label{sec:prompts}
\FloatBarrier
\subsection{Issue Agent}
\FloatBarrier

\begin{boxL}
You are a consultant hired by a large corporation. You will be given five negative reviews from customers. 
Your task is to identify the major themes, and list the specific issues the customers faced under that theme.
Do not include explanations or restatements. The issues should be descriptive, concise and clear. 

The output must strictly follow the json format.
Here is an example:
\\
\\
My reviews are:
"The house looked great in the photos, but once I moved in the reality was very different. There are cracks in the walls and baseboards, sloppy patchwork that looks like it was rushed, and the sprinkler system doesn't work at all. I was told I'd have to spend thousands to fix it, just a month after moving in. The windows fall off track, rain seeps in because the seals are broken, and even the fridge was missing the water and air filters."
\\
\\
"Honestly the worst service I've ever dealt with. Their employees hang up on you all the time, and when they don't, you're stuck waiting on hold for hours. I've spent so much time on the phone trying to get them to fix an issue they admitted was a mistake, but nothing ever gets resolved. Multiple people promised me a call back and, of course, no one ever did."
\begin{lstlisting}[language=json, numbers=none, xleftmargin=0pt]
{
"a": {
    "theme": "Maintenance",
    "issues": ["Sprinkler system damaged", "Windows falling off track", "Cracks in the wall"]
},
"b": {
    "theme": "Customer Support",
    "issues": ["Employees hang up on you constantly", "You have to wait for hours on the phone", "No call backs from support"]
}
}
\end{lstlisting}

Rules:\\
1) A theme can contain at most 5 issues.\\
2) The themes and the issues under them should not be repetitive.\\
3) If there are multiple issues that are similar to each other, merge them into one.\\
4) An issue can belong to at most one theme.
\end{boxL}

\subsection{Recommendation Agent}
\begin{boxL}
You are a consultant tasked with solving issues faced by customers. You will be given an issue, along with the broad theme associated with that issue.
\\
Give 3 to 4 actionable business recommendations suited for the following issue and theme:\\
\verb|<theme>| \\
\verb|<issue>| \\
You must only generate the recommendations, and not any other additional text, explanations, benefits.

\end{boxL}

\subsection{Evaluation Agent}
\begin{boxL}
Evaluate the advice: {advice} for the problem: {issue}. The issue is associated with the following theme: {theme}.
Provide scores from 1 (poor) to 5 (excellent) for Specificity, Relevance, Actionability and Concision.
If the advice is lacking, provide feedback on how to improve it. Here are examples of how to score:
\end{boxL}

The few shot examples given are:\\ 
\begin{lstlisting}[language=json, numbers=none, xleftmargin=0pt, frame= single, framerule=0.4pt]
[
  {
    "issue": "Long wait times at check-in (30-60+ minutes)",
    "advice": "Introduce a pre-arrival web check-in system that allows guests to upload their identification, confirm payment details, and choose their room preferences before arriving. This will reduce manual data entry at the front desk and speed up the process, especially during peak hours when lines tend to build up quickly.",
    "SRAC": [5, 5, 5, 4],
    "explanation": "Highly specific, relevant, and actionable advice that directly addresses the issue. Slightly long but precise and feasible for implementation."
  },
  {
    "issue": "Long wait times at check-in (30-60+ minutes)",
    "advice": "Encourage guests to arrive earlier in the day or during less busy times to avoid crowds. You could include this suggestion in their booking confirmation email or on the website so they can plan accordingly. This would help distribute guest arrivals throughout the day and reduce peak congestion.",
    "SRAC": [3, 4, 4, 4],
    "explanation": "Moderately specific and relevant. It provides a practical tip, though it doesn't fundamentally fix the process. Reasonably concise."
  },
  {
    "issue": "Long wait times at check-in (30-60+ minutes)",
    "advice": "Consider redesigning the entire hotel reception area to include a coffee bar, lounge seating, and entertainment screens. Guests could relax while waiting, which may make the wait seem shorter. It might also increase overall satisfaction and generate some additional revenue from the lobby cafe.",
    "SRAC": [3, 2, 2, 3],
    "explanation": "Creative but not very relevant or actionable in solving long wait times directly. Focuses more on perception than process improvement."
  },
  {
    "issue": "Long wait times at check-in (30-60+ minutes)",
    "advice": "Use AI-based forecasting tools to analyze booking data, flight schedules, and local events to predict check-in surges. Adjust staffing levels dynamically based on these forecasts and automate alerts for the management team to ensure resource allocation matches real-time demand.",
    "SRAC": [5, 5, 5, 4],
    "explanation": "Excellent strategic solution. Highly specific, directly relevant, and very actionable with the right tools. Slightly verbose but effective."
  },
  {
    "issue": "Long wait times at check-in (30-60+ minutes)",
    "advice": "Hire more employees to make check-in faster. The more people you have, the better and faster everything will be for guests. This should solve the issue without needing to change any systems or technology.",
    "SRAC": [2, 3, 3, 5],
    "explanation": "Very concise but oversimplified. Not specific about timing, training, or efficiency improvements. Lacks depth and sustainability."
  },
  {
    "issue": "Long wait times at check-in (30-60+ minutes)",
    "advice": "Install digital check-in kiosks at multiple points around the lobby where guests can verify their ID, pay deposits, and receive a digital key. This will streamline standard check-ins, leaving staff free to handle exceptions or special requests more quickly.",
    "SRAC": [5, 5, 4, 4],
    "explanation": "Clear, specific, and relevant. Actionable though requires upfront investment. Balanced in detail and brevity."
  },
  {
    "issue": "Long wait times at check-in (30-60+ minutes)",
    "advice": "Develop a multilingual mobile concierge chatbot to assist with pre-arrival check-in, explain policies, and collect preferences. The chatbot can guide guests through ID verification and notify reception staff before arrival to ensure all details are ready, cutting down on face-to-face processing time.",
    "SRAC": [5, 5, 5, 3],
    "explanation": "Very specific and forward-thinking. Highly actionable with good infrastructure, though slightly long and tech-heavy."
  },
  {
    "issue": "Long wait times at check-in (30-60+ minutes)",
    "advice": "Post motivational signs near the check-in counter reminding guests to stay patient and positive while they wait in line. You could also play calming background music to make the experience more pleasant for everyone.",
    "SRAC": [2, 2, 2, 4],
    "explanation": "Not specific or relevant to solving the root cause. Actionable but superficial, it improves perception, not efficiency."
  }
]
\end{lstlisting}

\subsection{Ranking Agent}
\begin{boxL}
You are a senior consultant hired by a corporation who wishes to improve customer satisfaction. You will be given three recommendations to solve an issue faced by the customer. You must decide which one of the three - first, second or third is the best option. Consider all aspects from the perspective of the corporation such as practicality, cost of implementation and efficacy in solving the issue.
\end{boxL}



\section{Outputs}
\begin{figure*}[h]
\centering
\includegraphics[width=1.0\textwidth]{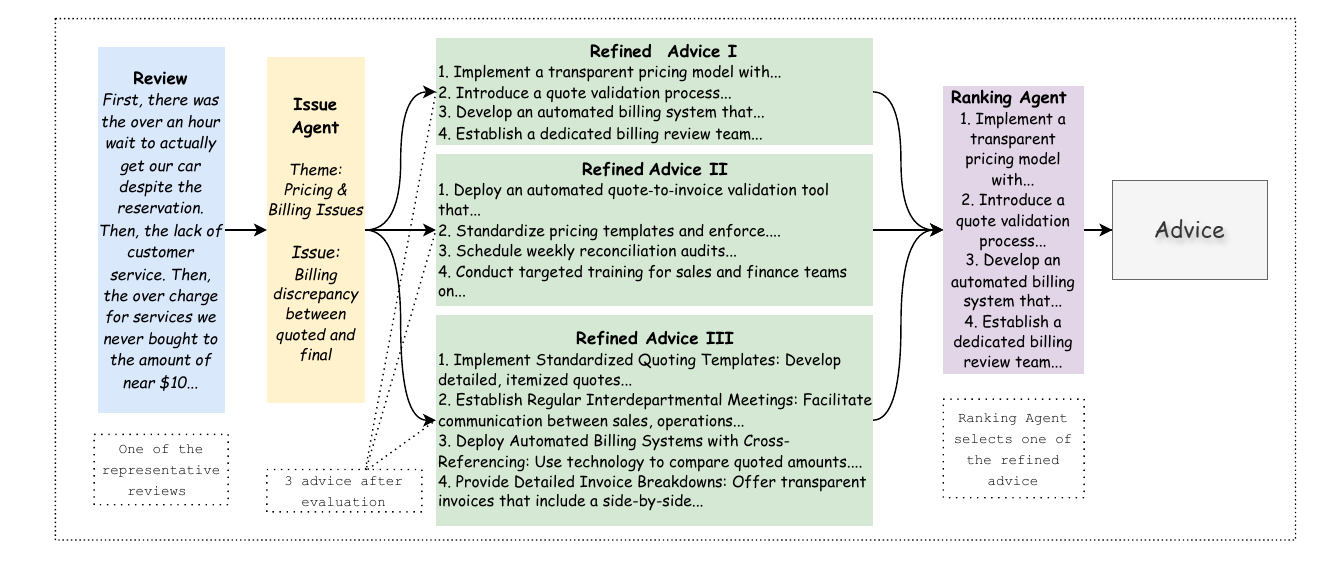}
\caption{\textbf{\textit{An example of the review-to-advice pipeline.}} A representative review may contain multiple issues (e.g., long wait times, lack of customer service and excessive charges). The Issue Agent extracts and maps all detected issues to one or more themes; for clarity, only one issue is shown in this figure for demonstration. For each extracted issue, multiple candidate advices are generated and iteratively refined, and the Ranking Agent selects the best refined advice as the final Advice(per issue).}
\label{fig:Workflow2}
\end{figure*}
\subsection{Output of the Issue Agent}
\label{sec:issue-output}
\subsubsection{Automotive}
\begin{lstlisting}[language=json, numbers=none, xleftmargin=0pt, frame= single, framerule=0.4pt]
{
    "a": {
        "theme": "Pricing & Billing Issues",
        "issues": [
            "Unexpected price increase at pickup",
            "Unauthorized insurance charges",
            "Extra gas charges",
            "Overcharge for services not requested",
            "Billing discrepancy between quoted and final"
        ]
    },
    "b": {
        "theme": "Customer Service Failures",
        "issues": [
            "Rude staff behavior",
            "Manager unresponsive or unavailable",
            "Failure to resolve complaints or provide callbacks",
            "Lack of assistance with bags/shuttle",
            "Poor communication during return process"
        ]
    },
    "c": {
        "theme": "Vehicle Condition",
        "issues": [
            "Vehicle had dents and scratches",
            "Interior was filthy",
            "Vehicle not ready at pickup"
        ]
    },
    "d": {
        "theme": "Shuttle & Pickup Issues",
        "issues": [
            "Off\u2011site pickup requiring shuttle",
            "Long wait for shuttle",
            "Shuttle driver rude and unhelpful"
        ]
    },
    "e": {
        "theme": "Transparency & Misrepresentation",
        "issues": [
            "Misrepresentation of prices",
            "Claims of mandatory insurance",
            "Lack of clarity on policies"
        ]
    }
}
\end{lstlisting}

\subsubsection{Restaurant}
\begin{lstlisting}[language=json, numbers=none, xleftmargin=0pt, frame= single, framerule=0.4pt]
{
    "a": {
        "theme": "Food Quality",
        "issues": [
            "Food served cold, stale, or not fresh",
            "Missing or incomplete dishes (e.g., no sauce, missing oysters)",
            "Food overpriced for quality",
            "Food served at incorrect temperature",
            "Food lacked proper seasoning"
        ]
    },
    "b": {
        "theme": "Service & Staff",
        "issues": [
            "Rude and unprofessional staff behavior",
            "Slow service and delayed responses",
            "Staff not checking on customers or attending to needs",
            "Staff discussing customers negatively",
            "Staff failed to remove or correct orders properly"
        ]
    },
    "c": {
        "theme": "Reservation & Wait Times",
        "issues": [
            "Reservation not honored or lost",
            "Long wait for seating or food"
        ]
    },
    "d": {
        "theme": "Pricing & Billing",
        "issues": [
            "Mandatory 20% service fee deemed excessive",
            "Billing errors (charged for unconsumed items, unclear discounts)"
        ]
    },
    "e": {
        "theme": "Kitchen Operations",
        "issues": [
            "Poor kitchen management leading to cold, stale, re-boiled dishes",
            "Lack of proper cooking (e.g., shrimp chewy, crab shell not crunchy)"
        ]
    }
}
\end{lstlisting}

\subsubsection{Hosptality}
\begin{lstlisting}[language=json, numbers=none, xleftmargin=0pt, frame= single, framerule=0.4pt]
{
    "a": {
        "theme": "Check\u2011in & Reservation",
        "issues": [
            "Long wait times at check\u2011in (30\u201360+ minutes)",
            "No greeting or assistance from staff upon arrival",
            "Overbooking confusion and inability to provide a second room",
            "Manager absent during issue resolution",
            "Delayed check\u2011in process (up to 3 hours)"
        ]
    },
    "b": {
        "theme": "Cleanliness & Maintenance",
        "issues": [
            "Unclean rooms with hair, snot, and general dirt",
            "Old carpet, bedding, wallpaper, and outdated d\u00e9cor",
            "Dirty bathrooms with hair, odor, and unclean surfaces",
            "Trash and unclean items in drawers and coffee maker",
            "Damaged or missing room features (cracked window, broken TV)"
        ]
    },
    "c": {
        "theme": "Noise & Room Location",
        "issues": [
            "Excessive noise from stairwell late at night",
            "Room located near stairwell causing disturbance"
        ]
    },
    "d": {
        "theme": "Staff Attitude & Service",
        "issues": [
            "Staff appear bored and unhelpful",
            "Dismissive response to complaints",
            "No follow\u2011up from housekeeping after reporting issues",
            "No call backs or assistance after complaints"
        ]
    },
    "e": {
        "theme": "Experience",
        "issues": [
            "Dealers replaced by machines",
            "Unfair machine payouts leading to significant losses",
            "Lack of crowd and excitement at gaming tables"
        ]
    },
    "f": {
        "theme": "Safety & Health",
        "issues": [
            "Bathroom accident resulting in hospitalization",
            "Slippery bathtub surfaces",
            "Smoke alarm baggies covering alarms",
            "No incident report provided after the accident"
        ]
    }
}
\end{lstlisting}

\subsection{Final Advice}
The following tables contain some examples of final advice for automotive, restaurant and hospitality industries respectively. These are the results from our proposed multi-agent system using three LLMs (Llama 3.3 70b Versatile, GPT OSS 120b, and Deepseek R1 Distill Llama 70b).  
\begin{table}[ht]
\centering
\small
\renewcommand{\arraystretch}{1.15}
\caption{Automotive (\textit{Payless Parking and Car Rental}) Dataset}
\begin{tabular}{@{}p{0.7cm}|p{3.2cm}|p{3.0cm}|p{7.6cm}@{}}
\toprule
S.N. & Issue & Theme & Advice \\
\midrule
1 & Extra gas charges & Pricing \& Billing Issues &
\begin{tabular}[t]{@{}l@{}}
\textbullet\ Implement a transparent gas charge calculation system on invoices\\
\textbullet\ Introduce a tiered pricing structure with clear gas charge thresholds\\
\textbullet\ Offer a gas charge estimator tool on the company website\\
\textbullet\ Provide a gas charge waiver option for customers who meet certain\\ criteria
\end{tabular}
\\
\midrule
2 & Failure to resolve complaints or provide callbacks & Customer Service Failures &
\begin{tabular}[t]{@{}l@{}}
\textbullet\ Implement a Complaint Resolution Tracker to monitor and follow\\ up on customer complaints.\\
\textbullet\ Introduce a Callback Guarantee, ensuring customers receive a response\\ within a specified timeframe.\\
\textbullet\ Develop a Customer Service Escalation Procedure to quickly address\\ and resolve complex issues.\\
\textbullet\ Provide Ongoing Training for Customer Service Representatives on\\ effective communication and issue resolution skills.
\end{tabular}
\\
\midrule
3 & Interior was filthy & Vehicle Condition &
\begin{tabular}[t]{@{}l@{}}
\textbullet\ Implement a mandatory vehicle inspection and cleaning protocol\\ before each rental.\\
\textbullet\ Provide customers with a vehicle condition report to sign off on at\\ the start of the rental period.\\
\textbullet\ Train staff on the importance of maintaining clean vehicles and\\ establish a system for reporting and addressing dirty vehicles.\\
\textbullet\ Offer a premium cleaning service for an additional fee to customers\\ who want an extra level of cleanliness.
\end{tabular}
\\
\midrule
4 & Long wait for shuttle & Shuttle \& Pickup Issues &
\begin{tabular}[t]{@{}l@{}}
\textbullet\ Implement a real-time shuttle tracking system to provide customers\\ with accurate estimated arrival times.\\
\textbullet\ Increase the frequency of shuttle runs during peak hours to reduce\\ wait times.\\
\textbullet\ Designate specific pickup points with clear signage and seating\\ areas to improve the overall waiting experience.\\
\textbullet\ Offer a mobile app for customers to request pickups and receive\\ updates on shuttle schedules and delays.
\end{tabular}
\\
\midrule
5 & Misrepresentation of prices & Transparency \& Misrepresentation &
\begin{tabular}[t]{@{}l@{}}
\textbullet\ Implement a clear and consistent pricing policy across all sales\\ channels.\\
\textbullet\ Provide detailed breakdowns of costs and fees associated with each\\ product or service.\\
\textbullet\ Introduce a price match guarantee to ensure customers are not\\ overcharged.\\
\textbullet\ Conduct regular audits to detect and prevent \\misrepresentation of prices by sales representatives or partners.
\end{tabular}
\\
\bottomrule
\end{tabular}
\end{table}

\begin{table}[ht]
\centering
\small
\renewcommand{\arraystretch}{1.15}
\caption{Restaurant (\textit{New Orleans Creole Cookery}) Dataset}
\begin{tabular}{@{}p{0.7cm}|p{3.2cm}|p{3.0cm}|p{7.6cm}@{}}
\toprule
S.N. & Issue & Theme & Advice \\
\midrule
1 & Food served cold, stale, or not fresh & Food Quality &
\begin{tabular}[t]{@{}l@{}}
\textbullet\ Implement a First-In-First-Out inventory system to ensure older\\ products are sold or used before newer ones.\\
\textbullet\ Conduct regular temperature checks on storage and display equipment\\ to prevent food from being stored at incorrect temperatures.\\
\textbullet\ Introduce a quality control checklist for kitchen staff to verify\\ the freshness and temperature of food before serving.\\
\textbullet\ Establish a supplier evaluation and rating system to ensure that\\ only high-quality, fresh ingredients are sourced.
\end{tabular}
\\
\midrule
2 & Rude and unprofessional staff behavior & Service \& Staff &
\begin{tabular}[t]{@{}l@{}}
\textbullet\ Implement a comprehensive customer service training program for\\ all staff members.\\
\textbullet\ Develop a clear code of conduct and disciplinary policy for \\unprofessional behavior.\\
\textbullet\ Introduce a staff evaluation and feedback system to monitor and\\ address rude behavior.\\
\textbullet\ Hire a dedicated customer service manager to oversee staff \\interactions and provide coaching.
\end{tabular}
\\
\midrule
3 & Reservation not honored or lost & Reservation \& Wait Times &
\begin{tabular}[t]{@{}l@{}}
\textbullet\ Implement a digital reservation system with automated confirmations\\ and reminders.\\
\textbullet\ Train staff to thoroughly verify reservations upon arrival and\\ have a clear protocol for handling lost or misplaced reservations.\\
\textbullet\ Offer a waitlist management system that provides real-time updates\\ to customers on wait times and reservation status.\\
\textbullet\ Establish a clear escalation procedure for customers whose reservations\\ are not honored, including a dedicated customer service \\contact and potential compensation or priority seating on the next\\ available table.
\end{tabular}
\\
\midrule
4 & Mandatory 20\% service fee deemed excessive & Pricing \& Billing &
\begin{tabular}[t]{@{}l@{}}
\textbullet\ Introduce a tiered service fee structure with varying percentages\\ based on the total bill amount.\\
\textbullet\ Offer an optional service fee waiver for loyal customers or those\\ who meet certain spending thresholds.\\
\textbullet\ Implement a transparent pricing model that clearly breaks down\\ the service fee and other charges.\\
\textbullet\ Consider reducing the mandatory service fee to 15\% and introduce\\ additional revenue streams through premium services.
\end{tabular}
\\
\midrule
5 & Lack of proper cooking & Kitchen Operations &
\begin{tabular}[t]{@{}l@{}}
\textbullet\ Implement a standardized cooking time and temperature control\\ system for all menu items.\\
\textbullet\ Provide ongoing training and certification programs for kitchen\\ staff on proper cooking techniques.\\
\textbullet\ Introduce a quality control checklist to ensure dishes meet culinary\\ standards before being served.\\
\textbullet\ Invest in kitchen equipment upgrades, such as precision thermometers\\ and cookers, to improve cooking consistency.
\end{tabular}
\\
\bottomrule
\end{tabular}
\end{table}

\begin{table}[ht]
\centering
\small
\renewcommand{\arraystretch}{1.15}
\caption{Hospitality (\textit{Circus Circus Reno}) Dataset}
\begin{tabular}{@{}p{0.7cm}|p{3.2cm}|p{3.0cm}|p{7.6cm}@{}}
\toprule
S.N. & Issue & Theme & Advice \\
\midrule
1 & Delayed check-in process (up to 3 hours) & Check-in \& Reservation &
\begin{tabular}[t]{@{}l@{}}
\textbullet\ Implement a digital check-in system to reduce wait times and\\ allow guests to complete the check-in process online before arrival.\\
\textbullet\ Increase front desk staffing during peak hours to handle a higher\\ volume of check-ins and reduce delays.\\
\textbullet\ Introduce a mobile check-in app that enables guests to check-in\\ remotely and receive a digital key, bypassing the need to wait in \\line at the front desk.\\
\textbullet\ Offer a priority check-in service for loyalty program members or\\ guests who have paid for an upgraded experience to minimize their \\wait time.
\end{tabular}
\\
\midrule
2 & Unclean rooms with hair, snot, and general dirt & Cleanliness \& Maintenance &
\begin{tabular}[t]{@{}l@{}}
\textbullet\ Implement a rigorous room inspection checklist for housekeeping\\ staff to ensure thorough cleaning of all areas.\\
\textbullet\ Increase the frequency of deep cleaning and sanitization of \\high-touch areas and surfaces.\\
\textbullet\ Provide ongoing training and feedback to housekeeping staff on\\ proper cleaning techniques and protocols.\\
\textbullet\ Install a quality control process to monitor and address guest\\ complaints regarding room cleanliness.
\end{tabular}
\\
\midrule
3 & Excessive noise from stairwell late at night & Noise \& Room Location &
\begin{tabular}[t]{@{}l@{}}
\textbullet\ Install soundproofing materials in the stairwell and adjacent rooms.\\
\textbullet\ Implement quiet hours policy with increased security patrols during\\ late nights.\\
\textbullet\ Relocate rooms affected by stairwell noise to quieter areas of the\\ building.\\
\textbullet\ Install noise-reducing doors and seals on rooms near the stairwell.
\end{tabular}
\\
\midrule
4 & Dismissive response to complaints & Staff Attitude \& Service &
\begin{tabular}[t]{@{}l@{}}
\textbullet\ Implement a comprehensive customer complaint handling training\\ program for all staff members.\\
\textbullet\ Introduce a mystery shopping initiative to monitor staff attitude\\ and service quality.\\
\textbullet\ Develop a clear and transparent complaint escalation procedure\\ to ensure timely resolution.\\
\textbullet\ Conduct regular staff feedback sessions to identify and address\\ underlying issues contributing to dismissive responses.
\end{tabular}
\\
\midrule
5 & Lack of crowd and excitement at gaming tables & Experience &
\begin{tabular}[t]{@{}l@{}}
\textbullet\ Introduce themed nights and events to attract a specific crowd\\ and create a lively atmosphere.\\
\textbullet\ Implement a rewards program that incentivizes players to visit\\ and play at the gaming tables during off-peak hours.\\
\textbullet\ Offer interactive games and tournaments that allow players to \\compete against each other and create a sense of community.\\
\textbullet\ Hire energetic and charismatic dealers to create a more engaging\\ and entertaining experience at the gaming tables.
\end{tabular}
\\
\bottomrule
\end{tabular}
\end{table}

\end{document}